\documentclass[12pt, comma]{article}

\usepackage[a4paper,margin=1in]{geometry}

\usepackage{setspace}
\usepackage{newtxtext,newtxmath}
\usepackage{microtype}

\usepackage{lineno}
\linenumbers

\usepackage[english]{babel}

\usepackage{amsmath}
\usepackage{graphicx}
\usepackage[colorlinks=true, allcolors=black]{hyperref}
\usepackage{fontawesome}
\usepackage[dvipsnames,table]{xcolor}
\usepackage{array}
\usepackage{booktabs}
\usepackage{algorithm}
\usepackage{algorithmicx}
\usepackage{algpseudocode}
\usepackage{tabularx}
\usepackage{marvosym}
\usepackage[super, comma, sort&compress, numbers]{natbib}
\usepackage{blindtext}
\usepackage{float}
\usepackage{capt-of}
\usepackage{needspace}
\nolinenumbers

\usepackage{parskip}
\makeatletter

\makeatletter
 
\floatname{algorithm}{Algorithm}

\usepackage{caption}
\title{\textbf{ChipOmni: A General-Purpose AI Agent \\for Electronic Design Automation}}
\author{Pengkun Wang$^{1,2 \dagger \textsuperscript{\Letter}}$, Rui Mao$^{1,2 \dagger}$, Zhe Zhao$^{3 \dagger}$, Dongting Xie$^{1,2}$, \\Haibin Wen$^{4}$,Song Xu$^{6}$, Di Wu$^{1,2}$, Xu Wang$^{1,2}$, Zhengyang Zhou$^{1,2}$, \\Binwu Wang$^{1,2}$, Xiaofeng Cao$^{5}$, Yang Wang$^{1,2}$\textsuperscript{\Letter}\\ \\
$^{1}$ University of Science and Technology of China, Hefei, China\\
$^{2}$ Suzhou Institute for Advanced Research, \\University of Science and Technology of China, Suzhou, China\\
$^{3}$ Stanford University, Stanford, CA, USA\\
$^{4}$ City University of Hong Kong, Hong Kong SAR, China \\
$^{5}$ Tongji University, Shanghai, China\\
$^{6}$ TeraPrime, Suzhou, China\\
\\
$\dagger$ Contributed equally: \\Pengkun Wang, Rui Mao, Zhe Zhao\\\\
\textsuperscript{\Letter} Corresponding author \textit{e}-mails:\\ \textit{pengkun@ustc.edu.cn} (P.W.); \textit{angyan@ustc.edu.cn} (Y.W.)
}

\date{}

\begin{document}
\maketitle
\newpage
\noindent\textbf{Abstract}

\noindent\textbf{Electronic design automation (EDA) workflows remain highly fragmented, requiring engineers to coordinate heterogeneous representations, specialised software tools, and iterative verification across multiple design stages. Although large language models have brought more general reasoning and code-generation capabilities to EDA, many existing approaches remain focused on isolated tasks and struggle to plan, verify, and repair designs autonomously within executable engineering environments. Here we introduce ChipOmni, a knowledge-enhanced, context-evolving AI agent framework for EDA. ChipOmni integrates multimodal understanding, dynamic planning, multi-agent collaboration, and verification-driven tool interaction into a unified autonomous system. By continually incorporating feedback from external engineering tools, the framework maintains an evolving task context, integrates domain knowledge, and iteratively refines intermediate designs. Across benchmarks for analog circuit, RTL, testbench, and formal-assertion generation, ChipOmni improves multiple task-specific measures over the evaluated baselines and solves all 30 AnalogCoder-Pro tasks. In a long-horizon case study, ChipOmni advances a natural-language RISC-V specification through three design levels to a micro-SoC GDSII layout that passes physical sign-off, without human intervention after initialization. Together, these results show that ChipOmni can coordinate verification-driven design across heterogeneous tasks and sustain long-horizon engineering workflows, advancing a shift from fragmented, tool-centred processes towards agent-centred, closed-loop systems.
}
\newpage

\section*{Main}
Electronic design automation (EDA) underpins modern semiconductor innovation by transforming increasingly complex specifications into the integrated circuits that power computing, communication, and artificial intelligence \cite{appenzeller2026accelerating}. Specialized algorithms and tools already automate many individual stages, including hardware-description generation, simulation, logic synthesis, formal verification, and physical design \cite{mirhoseini2021graph,iordanou2024low}. Yet constructing a complete design workflow still requires engineers to coordinate heterogeneous representations, including natural-language specifications, circuit diagrams, hardware-description languages, testbenches, netlists, and physical constraints, across tools that expose different interfaces and failure modes \cite{sanchez2023comprehensive}. As design cycles shorten and system complexity increases, this coordination burden slows iteration, raises engineering cost, and constrains the pace of semiconductor innovation \cite{appenzeller2026accelerating,sanchez2023comprehensive}.

Recent advances in artificial intelligence have begun to reshape scientific \cite{feng2026agentic,ghareeb2026multi,gottweis2026accelerating,ashokkumar2026large,ding2025scitoolagent,swanson2025virtual} and engineering workflows \cite{huang2026autonomous,ferber2026towards,lupoiu2025multi} by coupling general-purpose models with specialized computational tools. In EDA, machine-learning and large-language-model methods have been applied to tasks including hardware-description generation, circuit understanding, specification analysis, and design-space exploration \cite{xu2025large,kang2025fveval,skelic2025circuit,luo2026machine,karahan2024deep}. AnalogCoder uses training-free code generation for analog circuit design \cite{lai2025analogcoder}; LaMAGIC2 targets language-model-based analog topology generation \cite{chang2025lamagic2}; and LEGO develops a skill-based platform for RTL generation \cite{lou2026lego}. ChatEDA coordinates tool execution from RTL to GDSII \cite{wu2024chateda}, Design Conductor reports autonomous CPU design from a written specification to GDSII \cite{krishna2026design}, and FlowFixer-EDA uses tool feedback to diagnose and repair failures in RTL-to-GDS flows \cite{dong2026flowfixer}. At the infrastructure and evaluation levels, FluxEDA provides stateful interfaces for interaction with EDA tools \cite{chen2026fluxeda}, whereas FluxBench evaluates agents on tool-interactive RTL-to-GDS workflows \cite{deng2026can}. Together, these efforts advance different aspects of agentic EDA. \textit{A broader challenge is to coordinate multimodal inputs, analog and digital design, verification artefacts, and physical implementation within a shared executable environment, while retaining evolving constraints and using validation feedback throughout long-horizon workflows \cite{zhang2026fully}}.

An agent that spans these settings must meet three coupled requirements. First, it must interpret heterogeneous design inputs and representations while grounding its reasoning in an executable EDA environment that provides access to domain knowledge, software libraries, command-line tools, and external checkers. Second, it must evaluate candidate outputs against task-specific functional, formal, or physical criteria, using tool feedback to diagnose and repair failures rather than simply reject candidates. Third, it must preserve a structured engineering context as constraints, intermediate artefacts, and failure histories change over long trajectories. Together, these requirements call for a closed-loop framework in which planning, tool execution, verification and context updates inform one another.

Here we introduce ChipOmni, a knowledge-enhanced, context-evolving multi-agent framework for EDA that recasts heterogeneous design tasks as a verification-driven closed-loop process. ChipOmni grounds task planning in an executable EDA environment that connects domain knowledge with design artefacts, software libraries, command-line tools, and external checkers. Its knowledge-grounding layer retrieves domain principles, structural protocols and tool-specific constraints as reusable knowledge units, thereby constraining planning and generation by engineering requirements rather than by language priors alone. A planner decomposes a high-level objective into interdependent subtasks and coordinates specialized agents to generate, inspect, and modify the relevant artefacts. Execution agents interact with simulators, formal checkers, synthesis tools and other external evaluators, while their feedback is transformed into structured diagnoses that guide subsequent actions. ChipOmni further maintains an evolving engineering context that records task state, constraints, intermediate artefacts and debugging history, allowing information acquired in one iteration to inform later decisions. Rather than imposing a fixed workflow or optimizing a single artefact, ChipOmni dynamically selects and composes actions across specifications, circuit representations, RTL, verification collateral and implementation reports.

We evaluate ChipOmni across analog circuit design, RTL generation, self-checking testbench generation, formal-assertion generation, and multimodal engineering tasks. These experiments assess its performance across different design tasks and toolchains, while mechanism ablations and repair case studies examine the role of feedback-guided revision. We further assess its long-horizon execution in a three-level RISC-V design task. Starting from a natural-language specification, ChipOmni progresses from a minimal processor to a micro-SoC with a GDSII layout and completed physical sign-off, without human intervention after initialization. Together, these results position ChipOmni as a step beyond stage-specific EDA assistants, towards AI systems capable of coordinating heterogeneous tools, constraints and feedback across long-horizon engineering workflows.

\section*{Results}
\subsection*{ChipOmni framework}
Supporting multiple EDA tasks within a single framework is challenging because they differ in input and output representations, domain constraints, and tool interfaces. To address this challenge, we developed ChipOmni, a general-purpose AI agent framework for EDA that operates within an executable EDA environment (Fig. 1a). This environment connects multimodal design inputs, domain knowledge, and external engineering tools. Its knowledge base contains more than 10,000 textbooks and technical documents and over 20 EDA software manuals. It also provides access to more than 160 software libraries, over 30 command-line tools, and more than 40 reusable skills. Together, these resources cover more than 50 EDA tasks, including netlist, RTL and testbench generation, as well as formal verification. ChipOmni is therefore not confined to a single design representation or a fixed stage of the design flow.

The framework coordinates this environment through four interacting agents that couple execution, verification, repair and context evolution (Fig. 1b). The PlanExec agent translates mixed-modal inputs, including circuit diagrams, technical documentation and design intent, into executable plans, carries out design actions and collects observations from tool execution. It submits candidate artefacts to the Validator agent, which invokes external tools to evaluate their compliance with engineering constraints. Validator also derives stage-specific acceptance criteria from the task specification to assess both intermediate results and final deliverables. Validator returns feedback on failed checks to PlanExec, which passes execution or validation errors and their associated diagnostic information to the DebugMemory agent. DebugMemory analyses detailed execution traces within a focused local context, records diagnoses and formulates targeted repair tickets that guide PlanExec’s revisions. PlanExec further implements these revisions and resubmits the resulting artefacts to Validator. Meanwhile, the ContextOptimizer agent receives successful validation feedback from Validator and error-correction records from DebugMemory. It distils task-relevant knowledge from these records and incorporates it into the knowledge base. It also maintains a focused shared model context by compressing the execution history and retaining the constraints, observations and intermediate results that most relevant to the current stage. This context is further informed by prior knowledge from the knowledge base, allowing continually evolving experience to shape subsequent planning and execution. Overall, PlanExec and DebugMemory form the core action–reflection loop, while Validator supplies feedback from the external environment to guide progress on open-ended tasks. ContextOptimizer supports focused reasoning and consolidates experience, keeping accumulated knowledge traceable and the active context manageable as the execution history grows. The workflow initiated by PlanExec’s actions thus provides PlanExec with a more focused context for the current stage and richer guidance for subsequent actions, sustaining a closed-loop agent pipeline that enables the whole system to evolve by iterative execution over long-horizon tasks.

\begin{figure}[!htbp]
\centerline{\includegraphics[width=178mm]{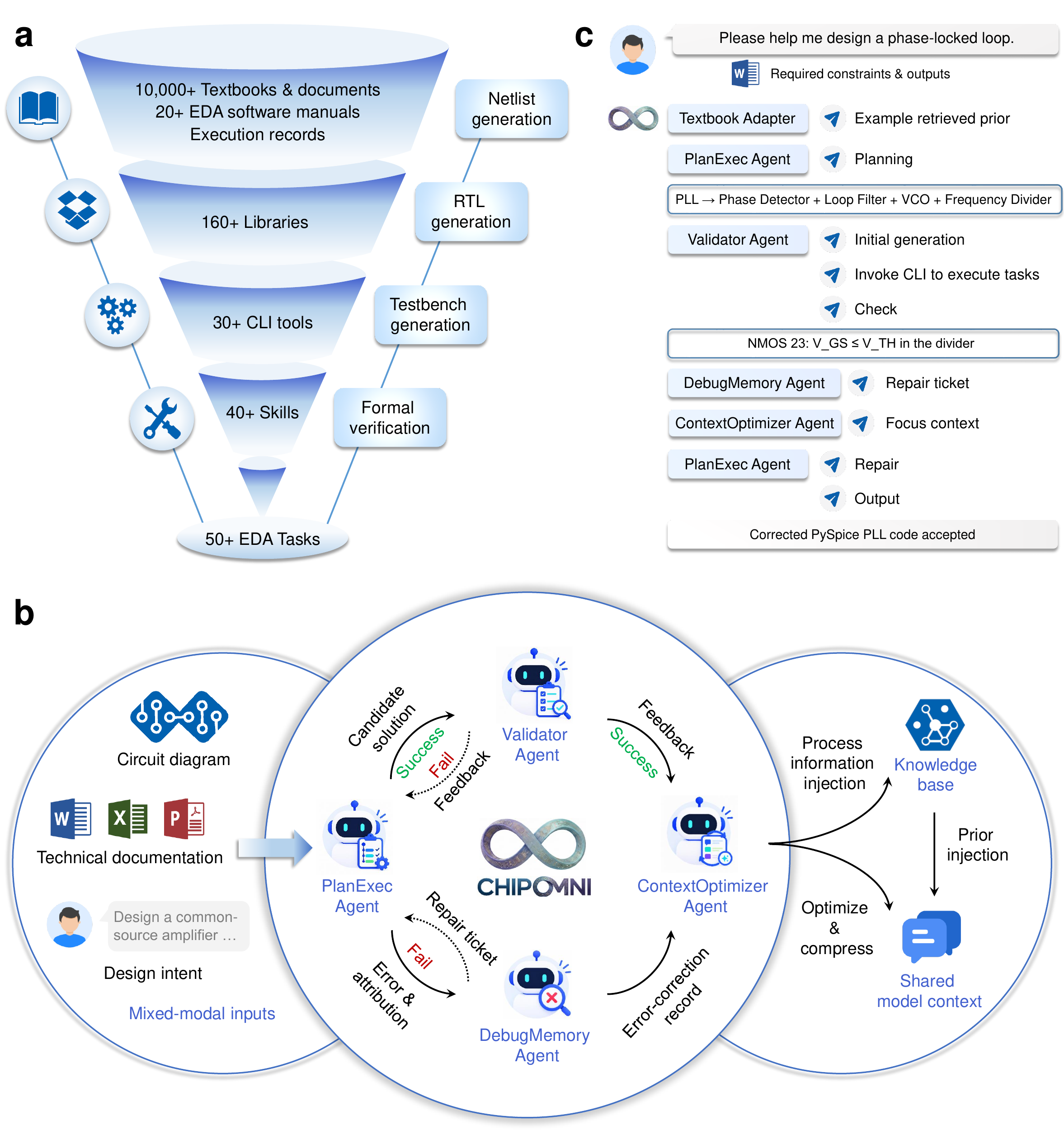}}
\caption{\label{fig:Figure1} \textbf{ChipOmni provides a knowledge-enhanced, closed-loop framework for heterogeneous EDA tasks.}  \textbf{a}, ChipOmni operates within an executable EDA environment that integrates textbooks and technical documents, EDA software manuals, execution records, software libraries, command-line tools, and reusable skills. These resources support more than 50 tasks. \textbf{b,} Mixed-modal design inputs are processed by four interacting agents. The PlanExec agent generates executable plans; the Validator agent invokes external tools and checks outputs; the DebugMemory agent stores diagnoses and repair tickets; and the ContextOptimizer agent manages process information, prior knowledge, and intermediate artefacts in a shared model context. \textbf{c,} Representative closed-loop trajectory for phase-locked-loop design. ChipOmni retrieves relevant prior knowledge, decomposes the design into circuit blocks, generates and executes PySpice code, detects a device-level constraint violation, and uses the resulting repair ticket and focused context to generate a corrected implementation that passes the configured validation.}
\end{figure}

Figure 1c illustrates this closed-loop process in a phase-locked-loop design task. From a natural-language request, the Textbook Adapter retrieved relevant prior knowledge, and the PlanExec agent decomposed the task into a phase detector, loop filter, voltage-controlled oscillator, and frequency divider. The Validator agent then executed the generated design through a command-line workflow and identified a device-level constraint violation in the divider. This failure was converted into a repair ticket, stored in DebugMemory, and incorporated into a focused context for the next planning cycle. The resulting repair produced a corrected PySpice implementation that passed the configured validation. Thus, ChipOmni connects generation, execution, validation, and repair within a single engineering trajectory, rather than treating them as isolated model calls.

\subsection*{General-purpose unification across heterogeneous EDA tasks}

To assess whether a single ChipOmni framework could operate across heterogeneous EDA tasks with distinct input modalities, target artefacts, validation procedures, and abstraction levels, we evaluated it across a suite of established benchmarks (Fig. 2a–i). The evaluation encompassed analog circuit topology generation, self-checking testbench generation, specification-to-RTL generation, hierarchical RTL generation, and specification-driven assertion generation. Across these tasks, ChipOmni generated executable engineering artefacts and incorporated task-specific feedback from circuit simulators, RTL compilers, simulation testbenches, and open-source and commercial formal-verification tools. This evaluation design enabled us to assess not only whether the generated artefacts were syntactically valid, but also whether they satisfied task-specific functional and verification criteria.

\begin{figure}[!htbp]
\centerline{\includegraphics[width=178mm]{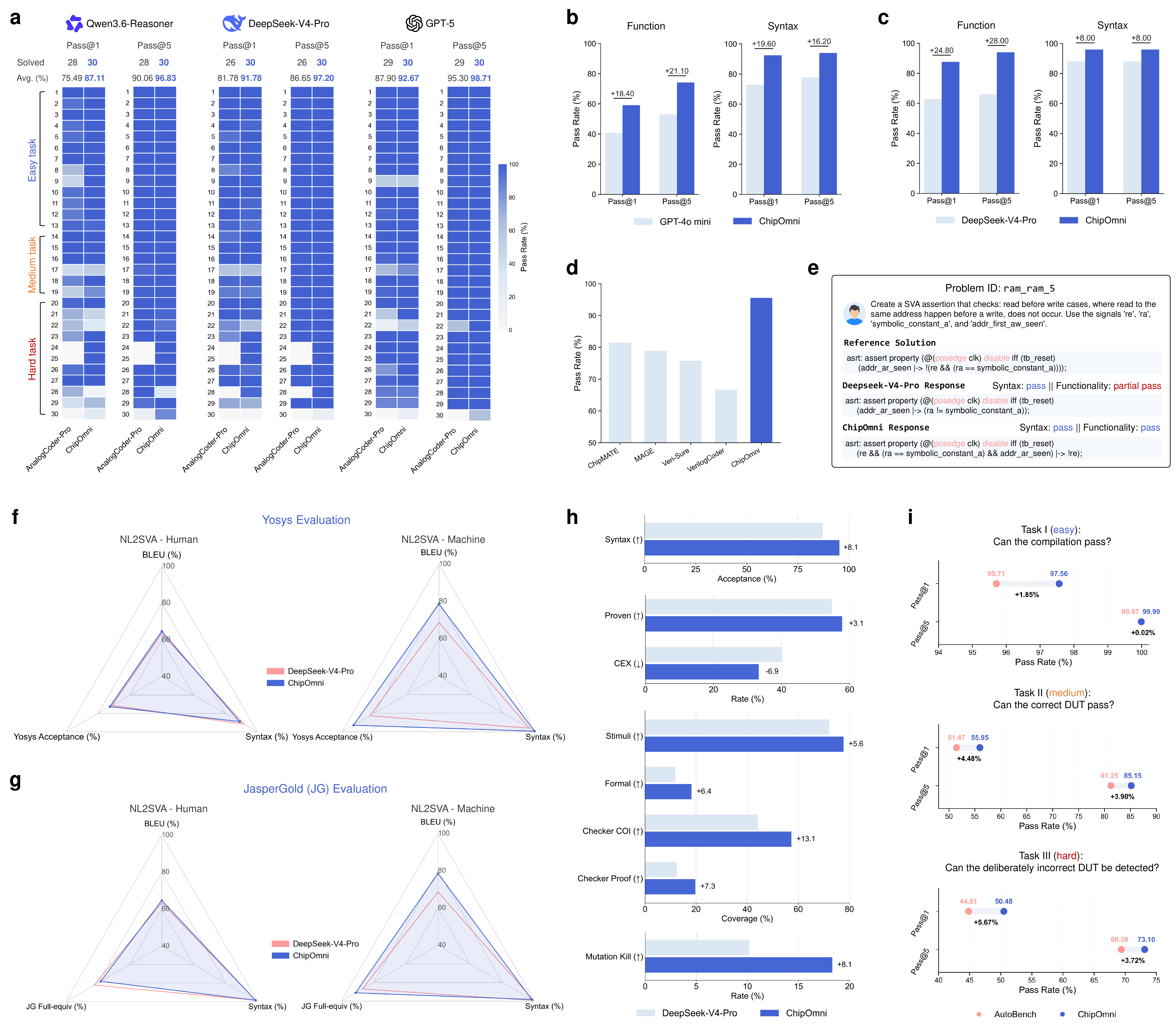}}
\caption{\label{fig:Figure2} \textbf{General-purpose performance of ChipOmni across heterogeneous EDA tasks and backbone models.} \textbf{a,} AnalogCoder-Pro evaluation of analog circuit generation using Qwen3.6-Reasoner, DeepSeek-V4-Pro and GPT-5. Heat maps show task-level Pass@1 and Pass@5 results; ChipOmni solved all 30 tasks with each backbone model. \textbf{b,c,} Functional and syntactic correctness on RTLLM2 using GPT-4o-mini (b) and DeepSeek-V4-Pro (c), respectively. ChipOmni improved both Pass@1 and Pass@5 relative to the corresponding backbone model. \textbf{d,} Pass@1 on VerilogEval v2 under the specification-to-RTL setting, where ChipOmni achieved the highest result among the evaluated methods. \textbf{e,} Representative FVEval example comparing the reference assertion, the DeepSeek-V4-Pro response and the ChipOmni response. \textbf{f,g,} NL2SVA-Human and NL2SVA-Machine results evaluated using the open-source Yosys flow (f) and JasperGold (g). Radar plots report BLEU, syntactic correctness, and Yosys acceptance or JasperGold full equivalence. \textbf{h,} Assertion generation on the 83-design AssertLLM2 benchmark using DeepSeek-V4-Pro as the backbone. ChipOmni improved syntax acceptance, proven rate, four coverage measures, and mutation kill rate, while reducing the counterexample rate. All metrics are higher-is-better except counterexample rate. \textbf{i,} Self-checking testbench generation on AutoBench using GPT-4o. ChipOmni outperformed AutoBench on testbench compilation (Eval0), acceptance of the correct RTL implementation (Eval1), and agreement with the golden testbench in detecting RTL mutants (Eval2), reaching Pass@1 of 97.56\% and Pass@5 of 99.99\% on Eval0.}
\end{figure}

We evaluated analog circuit generation on the 30-task AnalogCoder-Pro benchmark \cite{lai2025analogcoder,11432899} (Fig. 2a). The tasks comprised 8 easy, 5 medium, and 17 difficult designs spanning amplifiers, current mirrors, comparators, filters, operational amplifiers, oscillators, mixers, voltage-controlled oscillators, and phase-locked loops. We evaluated ChipOmni and the AnalogCoder-Pro baseline using Qwen3.6-Reasoner, DeepSeek-V4-Pro, and GPT-5. ChipOmni improved both Pass@1 and Pass@5 with each backbone. With Qwen3.6-Reasoner, Pass@1 increased from 75.49\% to 87.11\% and Pass@5 from 90.06\% to 96.83\%. The corresponding values increased from 81.78\% to 91.78\% and from 86.65\% to 97.20\% with DeepSeek-V4-Pro, and from 87.90\% to 92.67\% and from 95.30\% to 98.71\% with GPT-5. ChipOmni produced at least one successful solution for all 30 tasks with each backbone. Generated PySpice implementations were evaluated through netlist generation, operating-point analysis, and task-specific DC, AC, transient, or FFT simulations.

We next evaluated complete RTL generation across two benchmarks (Fig. 2b–d). RTLLM2 \cite{lu2024rtllm,liu2024openllm} comprises 50 hierarchical designs spanning arithmetic, control, memory, and other functional categories, including RISC-V components. With GPT-4o-mini as the backbone, ChipOmni improved functional Pass@1 and Pass@5 by 18.4 and 21.1 percentage points, respectively, while increasing the corresponding syntax pass rates by 19.6 and 16.2 percentage points (Fig. 2b). The gains were reproduced with DeepSeek-V4-Pro: functional Pass@1 and Pass@5 increased by 24.8 and 28.0 percentage points, respectively, and both syntax metrics increased by 8.0 percentage points (Fig. 2c). We further evaluated specification-to-RTL generation on the 156-problem VerilogEval v2 benchmark \cite{liu2023verilogeval,10.1145/3718088} using DeepSeek-V4-Pro as the backbone (Fig. 2d). Each output was compiled with Icarus Verilog and evaluated using the official testbench, with at most one feedback-based refinement step. ChipOmni achieved the highest Pass@1 among the evaluated methods, outperforming ChipMATE \cite{yu2026chipmate}, MAGE \cite{zhao2025mage}, VeriSure \cite{liu2026veri}, and VerilogCoder \cite{ho2025verilogcoder}. Together, these results show that ChipOmni consistently improved executable RTL generation across different backbone models, design descriptions, and levels of structural complexity.

Beyond RTL generation, we evaluated specification-driven SystemVerilog Assertion generation using FVEval \cite{kang2025fveval} (Fig. 2e–g). The analysis focused on 79 NL2SVA-Human instances, which pair expert-written natural-language properties with realistic testbenches, and 300 NL2SVA-Machine instances, which describe Boolean and temporal relationships using machine-generated language. Both settings provided three in-context examples and required a single assertion for each instance. A representative NL2SVA-Human case illustrates the distinction between syntactic validity and functional correctness: the DeepSeek-V4-Pro baseline produced an assertion that passed syntax checking but only partially captured the reference property, whereas ChipOmni generated an assertion that passed both syntax and functional evaluation (Fig. 2e).

We then evaluated the generated assertions using the open-source Yosys toolchain \cite{wolf2013yosys} (Fig. 2f). On NL2SVA-Human, ChipOmni increased BLEU from 63.4\% to 64.7\% and Yosys acceptance from 71.2\% to 72.6\%, although syntactic validity decreased from 91.8\% to 89.0\%. The gains were larger on NL2SVA-Machine, where BLEU increased from 68.7\% to 78.6\%, syntax from 96.1\% to 99.3\%, and Yosys acceptance from 82.7\% to 92.9\%. We independently assessed the same task families using the commercial JasperGold engine and its stricter full-equivalence criterion (Fig. 2g). On NL2SVA-Human, ChipOmni increased BLEU by 1.3 percentage points and preserved syntax at 98.6\%, but full equivalence decreased from 82.2\% to 78.1\%. On NL2SVA-Machine, BLEU increased by 9.9 percentage points and full equivalence increased from 86.9\% to 91.1\%, accompanied by a small reduction in syntax from 99.3\% to 98.2\%. Thus, ChipOmni produced consistent semantic gains on machine-generated property descriptions across both verification engines, whereas expert-written properties remained more challenging and exposed a trade-off between textual, syntactic, and strict equivalence metrics.

We further evaluated specification-driven assertion generation on the full AssertLLM2 benchmark \cite{wu2026assertllm2}, comprising 83 real-world RTL designs from 13 functional categories, including processors, communication controllers, cryptographic cores, signal-processing modules, memories, system controllers and systems-on-chip (Fig. 2h). Using DeepSeek-V4-Pro as the backbone model and one independent generation per design, ChipOmni was evaluated across syntax checking, formal-checking outcomes, design coverage and mutant detection. ChipOmni increased syntax acceptance from 87.2\% to 95.3\%. The proportion of assertions proven by formal verification increased from 54.7\% to 57.8\%, whereas the rate producing counterexamples decreased from 40.2\% to 33.3\%. Improvements extended across all four coverage measures: stimuli coverage increased from 72.1\% to 77.7\%, formal coverage from 11.8\% to 18.2\%, checker cone-of-influence coverage from 44.2\% to 57.3\%, and checker proof coverage from 12.4\% to 19.7\%. The mutation kill rate further increased from 10.2\% to 18.3\%. The concurrent gains in provability, coverage, and mutant detection indicate that ChipOmni improved not only the syntactic validity of generated assertions, but also their reach across the target designs and their ability to expose injected defects.

Finally, we evaluated self-checking testbench generation from natural-language descriptions and interface definitions using AutoBench \cite{qiu2024autobench}, which comprises 156 HDLBits tasks spanning 81 combinational and 75 sequential designs (Fig. 2i). We generated 10 candidate testbenches per task and allowed at most one feedback-based refinement. ChipOmni outperformed AutoBench across all three evaluation levels at both Pass@1 and Pass@5. For Eval0, which measures compilation success, ChipOmni increased Pass@1 from 95.71\% to 97.56\% and Pass@5 from 99.97\% to 99.99\%. For Eval1, which determines whether a generated testbench accepts the correct RTL implementation, Pass@1 increased from 51.47\% to 55.95\% and Pass@5 from 81.25\% to 85.15\%. For the more demanding Eval2 criterion, which measures agreement with the golden testbench when evaluating RTL mutants, ChipOmni improved Pass@1 from 44.81\% to 50.48\% and Pass@5 from 69.38\% to 73.10\%. These results show that the improvement extended beyond producing compilable testbenches to verifying correct designs and identifying deliberately faulty implementations.

Overall, these results show that ChipOmni can address EDA tasks with different input and output representations and validation procedures within a shared framework. Its improvements across multiple benchmarks demonstrate an ability to coordinate engineering tools, task constraints, and verification feedback across circuit domains and design stages.

\begin{figure}[!htbp]
\centerline{\includegraphics[width=178mm]{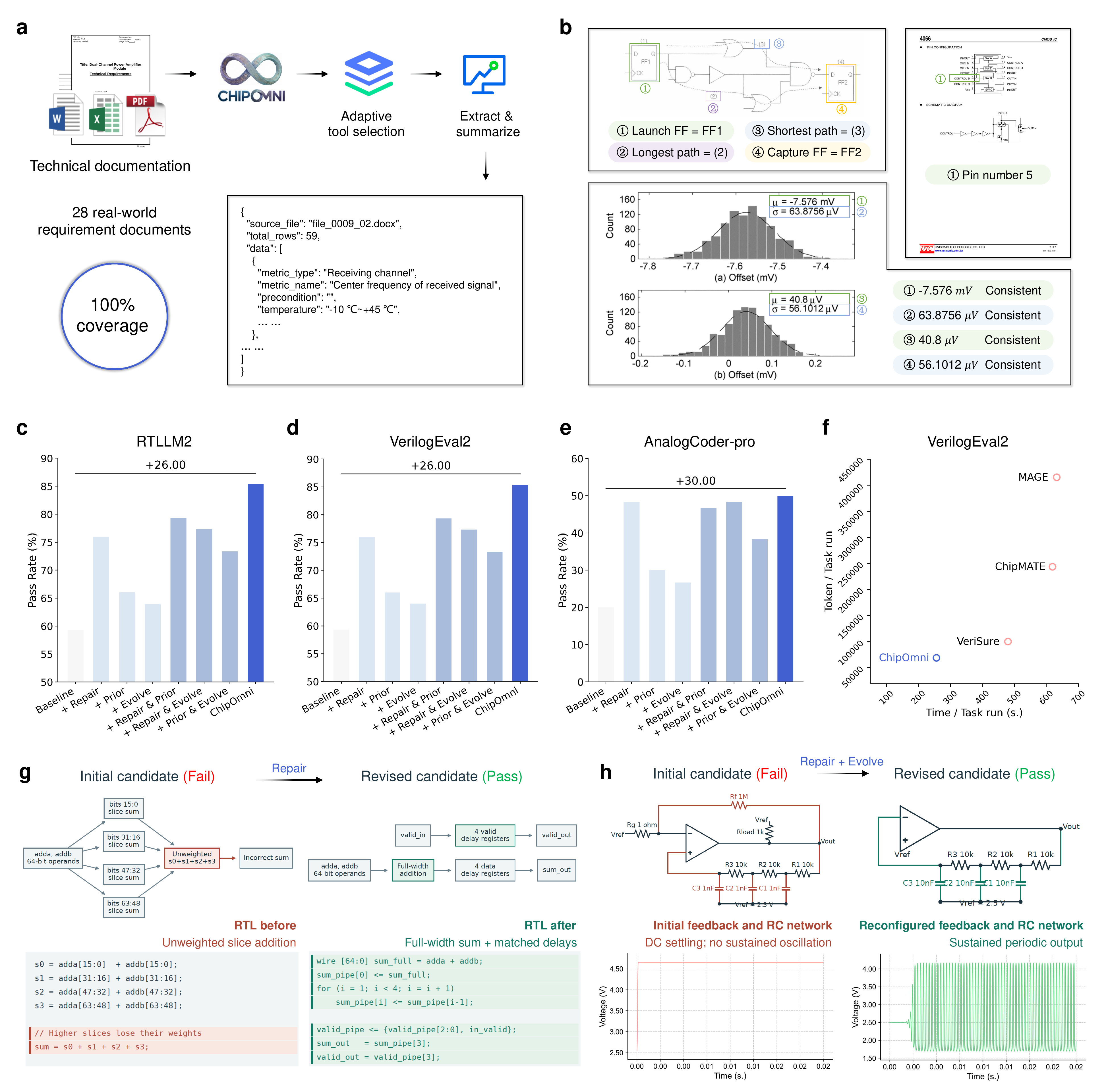}}
\caption{\label{fig:Figure3} \textbf{Multimodal reasoning, mechanism ablation and closed-loop execution in ChipOmni.} \textbf{a,} Adaptive extraction of requirements from Word, Excel and PDF documents into structured JSON. Across 28 real-world documents, ChipOmni covered 100\% of annotated requirement entries. \textbf{b,} Three MMCircuitEval examples using qwen3.8-chat as the backbone vision-language model: circuit-path identification, comparator offset-distribution analysis, and pin identification from a UTC 4066 datasheet. \textbf{c–e,} Pass-rate ablations on RTLLM2 (c), VerilogEval v2 (d), and AnalogCoder-Pro (e), comparing the baseline, individual mechanisms, all two-mechanism combinations, and the complete framework. Annotations show the complete framework's improvement over the baseline in percentage points. \textbf{f,} Runtime and token use per VerilogEval v2 task run for ChipOmni, MAGE, ChipMATE, and VeriSure. \textbf{g,} Digital repair replaces incorrect slice-wise addition with full-width addition and aligned data and valid delays. \textbf{h,} Analog Repair and Evolve reconfigure the feedback and RC network, changing a DC-settling output into sustained oscillation. The revised candidates pass verification.}
\end{figure}

\subsection*{Multimodal reasoning across heterogeneous engineering inputs}
ChipOmni converted design requirements distributed across Word, Excel, and PDF documents into structured records by adaptively selecting document-processing tools (Fig. 3a). The resulting JSON retained the source file and fields such as metric type, metric name, preconditions, and operating temperature. Across 28 real-world requirement documents, the extracted records covered 100\% of the annotated requirement entries. This workflow provided machine-readable requirements for subsequent design and verification tasks.

We further examined three representative MMCircuitEval \cite{zhao2025mmcircuiteval} cases using Qwen3.8-Chat as the backbone vision-language model (Fig. 3b). In a circuit diagram, ChipOmni combined the question’s cues with circuit connectivity to identify FF1 and FF2 as the launch and capture flip-flops, respectively, and paths (2) and (3) as the longest and shortest under the specified logic-depth criterion. From comparator offset-distribution plots, it extracted mean offsets of -7.576 $mV$ before compensation and 40.8 $\mu V$ afterwards, along with standard deviations of about 63.88 $\mu V$ and 56.10 $\mu V$. The result shows the mean shifting toward zero and a modest reduction in spread. In a datasheet pin diagram, ChipOmni matched the pin label to its number and identified pin 5 as CONTROL B on the UTC 4066. Together, these examples illustrate how the framework relates textual cues, numerical measurements, and visual connectivity to circuit-specific questions.

\subsection*{Ablation reveals the benefits of combining agentic mechanisms}
We assessed the contributions of Repair, Prior, and Evolve through module ablations on RTLLM2 \cite{wu2026assertllm2}, and representative subsets of VerilogEval v2 \cite{10.1145/3718088} and AnalogCoder-Pro \cite{11432899} (Fig. 3c–e). Each benchmark compared the baseline with single-module configurations, all three two-module combinations, and the complete ChipOmni framework. Repair produced the largest improvement among the single-module configurations on all three tasks. Combining all three mechanisms yielded the highest pass rate in each benchmark, exceeding the baseline by 26.00 percentage points on RTLLM2 and VerilogEval v2 and by 30.00 percentage points on AnalogCoder-Pro. On the analog tasks, this corresponded to an increase from 20\% to 50\%. The complete framework also outperformed every tested two-module configuration, showing that their joint use was more effective than any evaluated subset across these digital and analog tasks.

\subsection*{Efficient execution and verification-guided repair}
We compared runtime and token use per task run on VerilogEval v2 (Fig. 3f). Under the evaluated settings, ChipOmni required less time and fewer tokens than MAGE \cite{zhao2025mage}, ChipMATE \cite{yu2026chipmate}, and VeriSure \cite{liu2026veri}, occupying the lower-left region of the cost plot. We then examined how failing designs changed during repair. In the digital example, the initial candidate divided two 64-bit operands into four 16-bit slices and summed the slice-wise results without preserving the positional weights of the higher-order slices, producing an incorrect result (Fig. 3g). Repair replaced this operation with full-width addition, retained the carry-out in a 65-bit result, and introduced four-stage delay registers for both data and valid signals. The revised design aligned the result with its validity indicator and passed verification. The analog example showed a different failure mode (Fig. 3h). The initial op-amp feedback and three-stage RC network produced an output that settled to a DC level rather than sustained oscillation. Through Repair and Evolve, ChipOmni reconfigured the feedback and RC connections and increased the capacitance of each stage from 1 nF to 10 nF. The revised circuit produced a sustained periodic output after startup and passed verification.

\begin{figure}[!htbp]
\centerline{\includegraphics[width=178mm]{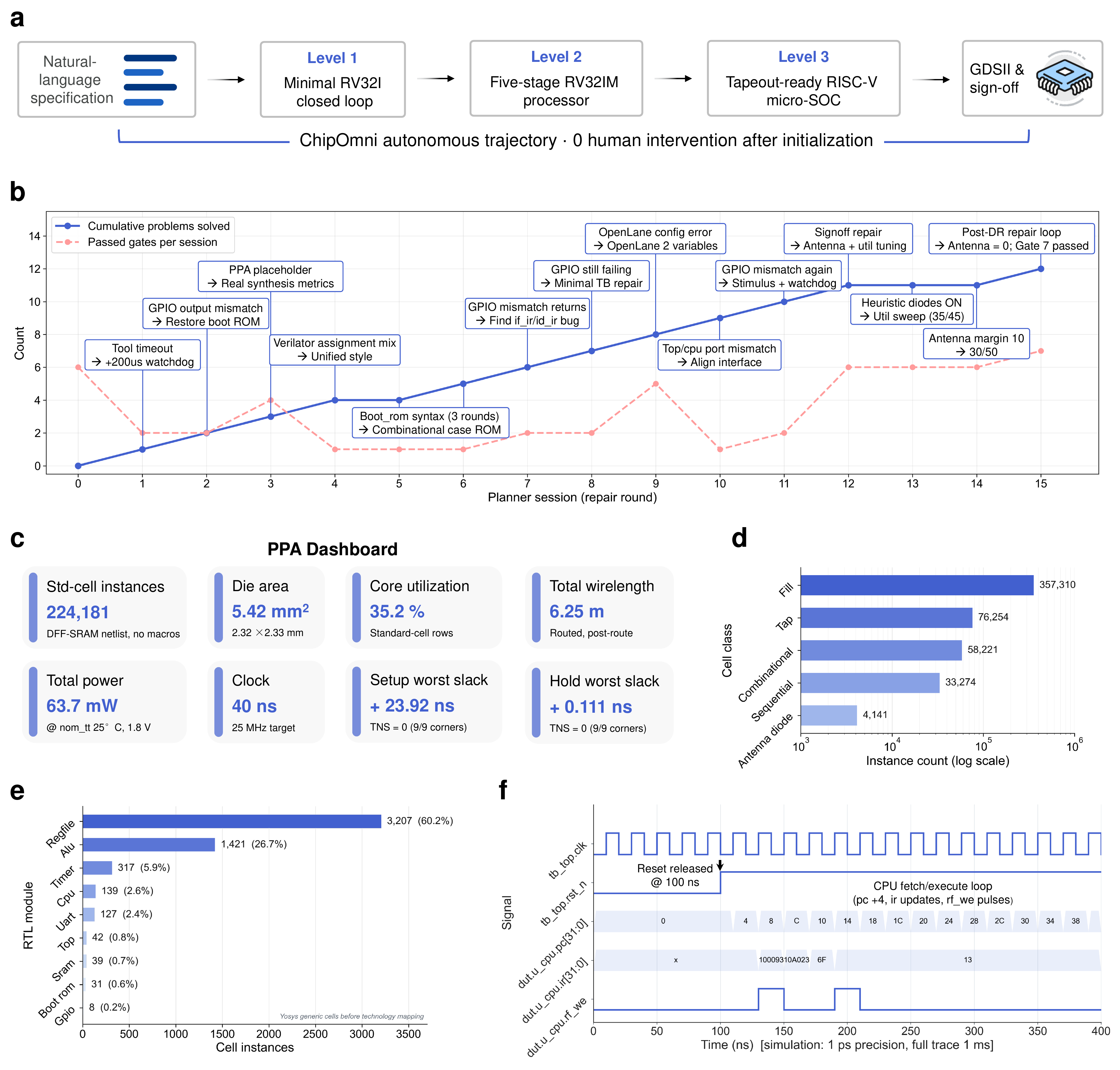}}
\caption{\label{fig:Figure4} \textbf{Autonomous execution of a long-horizon design task.} \textbf{a,} Three-level progression from a natural-language specification to a RISC-V micro-SoC, GDSII, and sign-off, without human intervention after initialization. \textbf{b,} Repair trajectory across 16 planner sessions. Blue shows cumulative resolved problems; pink shows evaluation gates passed per session. The final session records 12 resolved problems and seven passed gates. \textbf{c,} Final implementation and PPA metrics, including 224,181 non-filler instances, 5.42 mm² die area, and 63.7 mW analysed power. Setup and hold total negative slack are zero across nine corners at the 25-MHz target. \textbf{d,} Physical cell-class counts on a logarithmic scale, with filler cells counted separately. \textbf{e,} Yosys generic-cell distribution by RTL module before technology mapping. \textbf{f,} Representative RTL simulation showing reset release at 100 ns followed by instruction-fetch and register-file write activity.}
\end{figure}

\subsection*{Autonomous execution of a long-horizon design task}
To test whether ChipOmni could sustain engineering decisions across a long-horizon sequence of dependent EDA steps, we constructed a three-level RISC-V design task (Fig. 4a). Level 1 required a minimal RV32I processor, a self-checking testbench, and executable simulation and synthesis flows. Level 2 extended the design to a five-stage RV32IM processor with forwarding, load-use stalls, branch flushing, multiplication and division, and basic CSR support. Level 3 integrated the processor with boot ROM, SRAM, and memory-mapped UART, GPIO, timer, and SPI peripherals, then extended the task through physical implementation to GDSII generation and sign-off. Starting from a natural-language specification, ChipOmni progressed through these levels without human intervention after task initialization.

Figure 4b follows ChipOmni's repair decisions across the initial session and 15 subsequent planner sessions. Early feedback exposed execution and design-validity problems, including an external-tool timeout, boot-ROM syntax errors, and inconsistent RTL assignments. Recurrent GPIO mismatches then prompted changes to the boot program, testbench, pipeline logic, and CPU interface. In later sessions, ChipOmni adapted the OpenLane \cite{shalan2020building} configuration and adjusted antenna repair and utilization settings for physical sign-off. By the final session, it had resolved 12 recorded problems and passed all seven evaluation gates.

The final physical implementation contained 224,181 non-filler standard-cell instances and occupied a die area of 5.42 $mm^2$ (Fig. 4c). EDA power analysis yielded 63.7 $mW$ at the TT corner, 25 $^{\circ}\mathrm{C}$ and 1.8 $V$. At the 25-MHz target clock, static timing analysis gave worst setup and hold slacks of +23.92 $ns$ and +0.111 $ns$, respectively; total negative slack was zero for both checks across all nine analysed corners. We further examined the composition of the generated design. The physical layout contained 357,310 filler cells in addition to its non-filler instances (Fig. 4d). In the pre-mapping Yosys \cite{wolf2013yosys} analysis, the register file and arithmetic logic unit accounted for 60.2\% and 26.7\% of generic cells, respectively (Fig. 4e). A representative RTL simulation trace shows the generated processor operating after reset (Fig. 4f). Following reset deassertion at 100 $ns$, the program counter advanced through successive word addresses, the instruction register updated, and register-file write-enable pulses appeared. The trace captures instruction-fetch and execution activity in the illustrated simulation.

\begin{figure}[!t]
\centerline{\includegraphics[width=178mm]{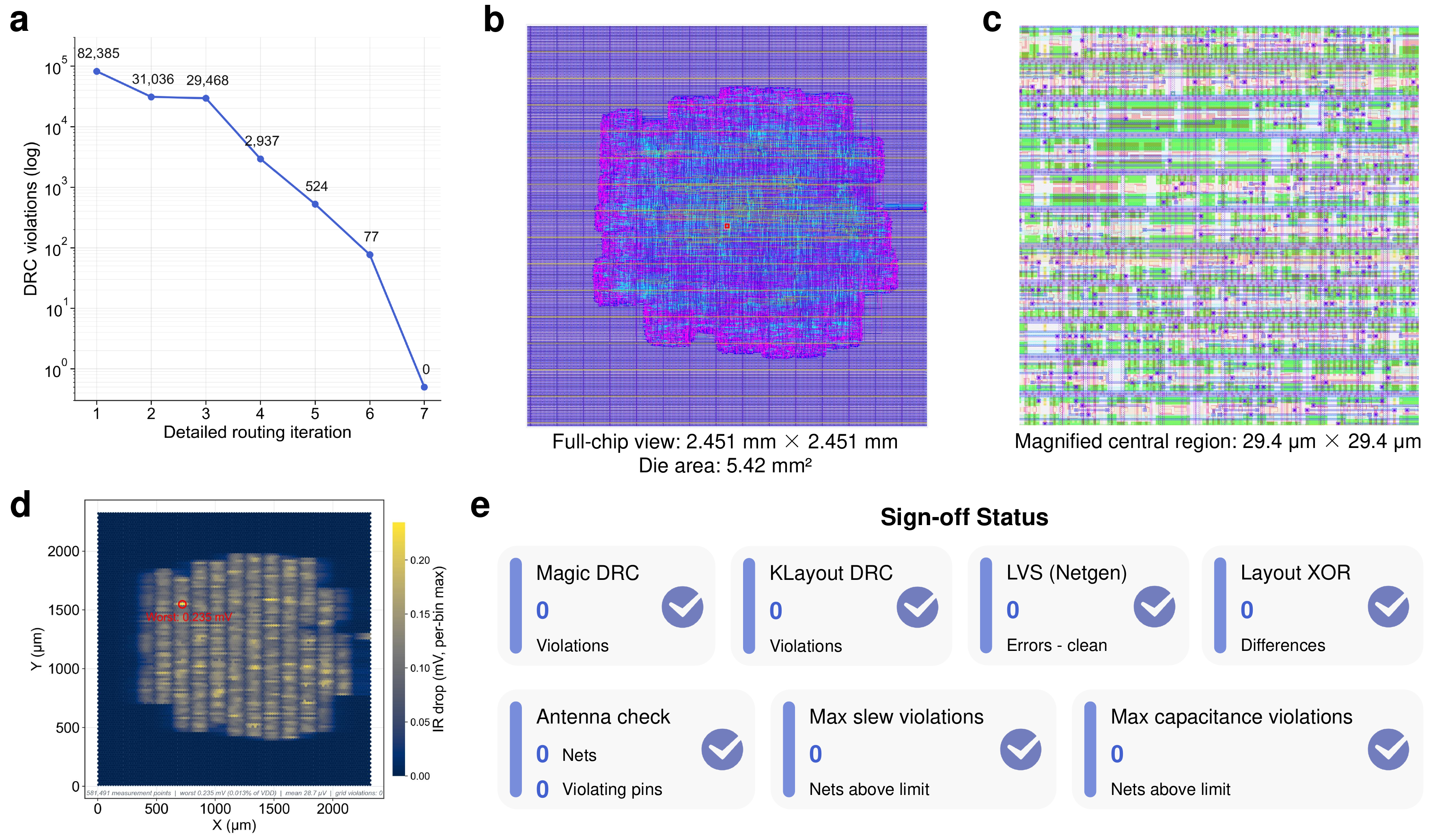}}
\caption{\label{fig:Figure5} \textbf{Routing convergence, physical layout, and sign-off of the RISC-V micro-SoC.} \textbf{a,} Detailed-routing violations decreased from 82,385 to zero over seven OpenROAD iterations; the zero endpoint is shown schematically on the logarithmic axis. \textbf{b,} Full-chip GDSII layout ($2,451 \mu m \times 2,451 \mu m$ field of view). \textbf{c,} Magnified view of the central region marked in b ($29.4 \mu m \times 29.4 \mu m$). \textbf{d,} Power-grid IR-drop distribution. The red marker indicates the maximum drop of 0.235 $mV$. \textbf{e,} Sign-off summary showing zero DRC violations, LVS errors, and layout XOR differences, with antenna, maximum-slew, and maximum-capacitance checks passed.}
\end{figure}

We next examined whether the autonomous design trajectory reached a physically verified layout. In a ChipOmni-orchestrated OpenROAD \cite{ajayi2019toward} run, seven detailed-routing iterations reduced design-rule violations from 82,385 to zero, achieving routing closure (Fig. 5a).

The final GDSII layout makes the scale of the autonomous design task visible (Fig. 5b). It implements the 5.42 $mm^2$ micro-SoC containing 224,181 non-filler standard-cell instances, carrying the design from a natural-language specification to a fully routed chip layout. The full-chip view shows the core, its interconnects, and the power-distribution structure. In the central region marked by the red box, a $29.4 \mu m \times 29.4 \mu m$ magnified view resolves orderly standard-cell rows and local multilayer routing (Fig. 5c). Power-grid analysis showed limited voltage drop across the implemented design (Fig. 5d). Across approximately 580,000 sampling points, the mean IR drop was 28.7 $\mu V$; the maximum was 0.235 $mV$, approximately 0.013\% of the 1.8-V supply. At physical sign-off, Magic and KLayout found no design-rule violations, Netgen found no layout-versus-schematic errors, and layout XOR found no differences. Antenna, maximum-slew, and maximum-capacitance checks also passed (Fig. 5e). Together with timing closure, these results show that ChipOmni completed the long-horizon task from a natural-language specification to routed GDSII and physical sign-off, without human intervention after initialization.

\section*{Discussion}

General-purpose EDA requires a decision process that remains useful when design representations, tools, and success criteria change. ChipOmni applied a shared planning, tool-use, and revision process to analog circuits, RTL, testbenches, and formal assertions, while evaluating each output through task-specific executable checks. The benchmark results suggest that heterogeneous design and verification tasks can be coordinated without imposing a single validation procedure on them.

The ablations clarify the contributions of individual mechanisms, while the repair cases show how validation feedback guides design revisions. Repair produced the largest gain among individual mechanisms, and the complete framework outperformed every tested subset. The three-level RISC-V task further tested whether this process could persist across dependent design stages. Starting from a natural-language specification, ChipOmni addressed problems in RTL, testbenches, tool configuration, and physical implementation without human intervention after initialization, ultimately producing a micro-SoC GDSII layout that passed physical sign-off. In a separate VerilogEval v2 comparison, ChipOmni used less time and fewer tokens per task run than the three compared methods, making execution cost another relevant measure of autonomous EDA systems.

Despite these advances, several questions remain in extending verification-driven autonomy to broader EDA workflows. ChipOmni’s planning and repair loop depends on accessible engineering tools, validation procedures that check key design constraints, and execution feedback sufficient to support diagnosis. Its applicability to requirements that are difficult to translate into executable checks, or to tasks with incomplete tool feedback, remains to be examined. Long-horizon execution without human intervention also entails sustained access to design files and external tools. Permission boundaries, safeguards against unintended actions, and execution auditing have not yet been systematically evaluated. Moreover, although the micro-SoC reached GDSII and passed physical sign-off, it has not yet been fabricated, and further silicon testing is needed to validate its functionality and performance.

The continued development of ChipOmni will focus on deeper verification and more reliable autonomous execution. Beyond using tool feedback to repair designs, we aim to assess the coverage of existing checks more systematically, identify design requirements that remain insufficiently verified, and plan additional simulation, formal analysis, or physical validation. We will also investigate how permission controls, auditable tool use, and failure recovery can be incorporated into execution, supporting more robust operation in practical engineering environments.

ChipOmni is being trialled with industry partners, and feedback from real design settings will inform its continued development. Building on its general-purpose ability to coordinate tasks, we plan to develop more specialised, high-performance agents for specific EDA functions and extend the framework to adjacent tasks. Future fabrication and silicon testing will provide a further assessment of the generated designs and bring measured outcomes into the design and verification loop. By combining a general-purpose framework with specialised capabilities and practical engineering feedback, ChipOmni could help move electronic design from fragmented tool automation towards autonomous, closed-loop engineering spanning specification, implementation and physical validation.

We plan to release the ChipOmni code repository and make its evaluation platform publicly available in a future release to support broader independent evaluation and further research.

\clearpage
\newpage

\bibliographystyle{naturemag}
\bibliography{main}
\end{document}